%% file: PaperOne.tex
\documentclass{article}

\usepackage{PRIMEarxiv}

\usepackage[utf8]{inputenc} 
\usepackage[T1]{fontenc}    
\usepackage{hyperref}       
\usepackage{url}            
\usepackage{booktabs}       
\usepackage{amsfonts}       
\usepackage{amsmath}        
\usepackage{nicefrac}       
\usepackage{microtype}      
\usepackage{lipsum}
\usepackage{fancyhdr}       
\usepackage{graphicx}       
\usepackage{subcaption}
\graphicspath{{media/}}     
\usepackage{placeins}   
\usepackage{float}      
\usepackage{multirow}
\usepackage{silence}
\usepackage{siunitx}
\hypersetup{
	pdftitle={On the Structural Limitations of Weight-Based Neural Adaptation and the Role of Reversible Behavioral Learning},
	pdfauthor={Pardhu Sri Rushi Varma Konduru},
	pdfsubject={Neural Adaptation, Structural Reversibility, RLAE},
	pdfkeywords={Neural Networks, Fine-Tuning, Catastrophic Forgetting, Reversibility, Parameter Isolation}
}

\title{On the Structural Limitations of Weight-Based Neural Adaptation and the Role of Reversible Behavioral Learning}

\author{
	Pardhu Sri Rushi Varma Konduru \\
	Malla Reddy University \\
	Hyderabad, Telangana, India.\\
	\texttt{pardhuvarma.cs@gmail.com} \\
}

\begin{document}
	\maketitle
	\thispagestyle{empty}

\begin{abstract}
	Neural models usually adapted and involve changes in parameters shared among the model components through fine-tuning, alignment-based training, and reinforcement learning. These changes have been found effective in short-term optimization. However, they result in long-term changes in the model’s basic behavior. In our study, we introduce the concept of structural irreversibility as a characteristic of shared-parameter model adaptation. This concept refers to the intertwining of task-specific objectives and representations of the model identity. We show that when parameters are directly mutated, the resulting model behaves divergently from the original model. This divergence cannot be reversed deterministically without an explicit snapshot of the parameters. We introduce reversible behavioral learning, in which model behaviors are structurally disassociated from identity parameters and can be deterministically unloaded through an explicit unload process. We also introduce the Recoverability Factor as a normalized measure of behavioral recoverability. We provide additional diagnostics based on model divergence. Experiments show that reversible model adaptation achieves rollback within numerical precision, whereas shared-parameter mutation displays persistent post-reset divergence.
\end{abstract}
	
	\section{Introduction}
	\label{sec:intro}
	For large neural models, adaptation typically involves direct updates to shared parameters. But this gives rise to a structural question: under what circumstances can behavioral change following adaptation be deterministically reversed without access to a dedicated parameter checkpoint? This paper examines the structural properties of neural adaptation mechanisms under sequential learning. We focus on the relationship between parameter modification, behavioral stability, and recoverability, and outline our empirical findings and contributions below.
	
	\noindent\textbf{Code Availability:} 
	The implementation and experiments supporting this work are publicly available. \cite{Varma2026rlae}.
	
	\subsection{Motivation} 
	\label{sec:motivation}
	After training, many complex neural models need to be adjusted to meet new tasks, safety, and what the model is intended to do in task specific scenarios. How these adjustments are made include adjusting the parameters over time, learning from human feedback, and other actions taken after training that impact the shared parts of the model.
	
	This adjustment process induces representational drift within the shared parameter space. Because the same parameters encode multiple abstractions, new updates may perturb representations supporting prior capabilities, leading to degraded reasoning, capability erosion, or unintended behavioral shifts.
	
	Also, our current capabilities are not very useful when we want to go back to the model’s previous state. There isn’t a simple way to go back to the model’s previous state using the techniques we have at hand. To reverse the previous actions, we either go back to a checkpoint or we have to retrain the model from scratch. This can be expensive, time-consuming, and it will hit the same areas of the model again.
	
	\subsection{Core Observation}
	\label{sec:observation}
	
	The key observation is that the reversibility of adaptation critically depends upon the representational location in which behavioral modification is represented. That is, when shared parameters are updated with adaptive changes, the resulting behavioral changes are represented over a space of parameters that simultaneously represent multiple abstractions. This means that task-specific goals and underlying identity representations are conflated during shared-parameter modification, a fact well-documented in the continual learning literature \cite{kirkpatrick2017ewc,delange2021continual}. Accordingly, deterministic reversal of adaptation is not possible, and the process of restoring behavior corresponds to an ill-posed problem without access to the original parameters \cite{kaushik2021rmn}.
	
	When the adaptive behavior is separated from the base model through the use of separable parameter components, such that the essential parameters remain static, improvements in retention and recoverability have been shown in the past \cite{kaushik2021rmn,lanzillotta2023isolation}. This is because the parameters that define the identity of the model remain unchanged, and hence, there is no need for inverse optimization, retraining, or checkpointing during the process of rollbacks, since the adaptive behavior can be simply uninstalled by removing the adaptive component.
	
	Crucially, the difference in this case demonstrates that reversibility is not primarily a matter of improved training procedures, more effective optimization, or stronger regularization. Rather, it is a property of the adaptation paradigm itself, namely whether or not the adaptive behavior remains coupled to or decoupled from the central representational spine. From this perspective, reversibility is not a curiosity of the training process but a consequence of shared-parameter adaptation, as has been observed in continual learning and parameter decoupling methods. \cite{delange2021continual,lanzillotta2023isolation}.
	
	\subsection{Contributions}
	\label{sect:contribution}
	
	This work makes the following contributions:
	\begin{itemize}
		\item We formalize a distinction between model identity and adaptive behavior in Section~\ref{sec:model_decomposition}, enabling precise reasoning about rollback and behavioral preservation.
		\item We identify \emph{structural irreversibility} in Section~\ref{sec:irreversibility}, as a fundamental limitation of weight-based neural adaptation, arising from the entanglement of task-specific objectives with shared model parameters.
		\item We empirically compare irreversible weight-based adaptation with reversible behavioral adaptation, demonstrating stark differences in post-reset recoverability.
		\item We formalize reversible behavioral adaptation through the notion of \emph{Runtime Low-Rank Adaptive Environments} (RLAE) in Section~\ref{sec:rlae}, where adaptive behavior is encoded in removable, runtime-controlled parameterizations while the base model remains frozen.
		\item We introduce \emph{recoverability} as an explicit evaluation criterion for adaptive neural systems in Section~\ref{sec:kl_rf}.
		\item We propose \emph{Structural Variance Analysis for Robustness} (SVAR) in Section~\ref{sec:svar}, as a diagnostic methodology for assessing behavioral stability under controlled perturbations.
	\end{itemize}

	\section{Background and Related Work}
	\label{sec:related_work}
	
	\subsection{Weight-Based Neural Adaptation}
	\label{sec:weight_based_adaptation}
	
	Weight-based neural adaptation refers to adaptation procedures in which behavioral change is realized through direct updates to a model’s parameter vector. Given a pretrained model parameterized by a shared parameter set (denoted abstractly as $\theta$), adaptation is performed by applying gradient-based optimization to modify $\theta$ in order to minimize a task- or objective-specific loss. This formulation underlies standard fine-tuning procedures, reinforcement learning from human feedback (RLHF), and many continual learning approaches \cite{kirkpatrick2017ewc,delange2021continual}.
	
	In this setting, the same set of parameters is shared across all objectives, and adaptation steps are informed by optimization and regularization techniques applied to the same representational foundations. As a result, weight-based adaptation becomes the standard adaptation technique, compared to other adaptations, such as isolating parameters or modularizing behavior, are compared.
	
	\subsection{Catastrophic Forgetting and Representation Drift}
	\label{sec:forgetting_drift}
	
	One of the important challenges in sequential weight-based learning is \emph{catastrophic forgetting}, which means that while we are learning new tasks, our performance on previous tasks keeps deteriorating. This is consistent with the stability-plasticity dilemma, where we need to keep our model plastic enough to learn new things but stable enough to retain what we already know \cite{kirkpatrick2017ewc,delange2021continual}.
	
	From previous research, it has been observed that forgetting is strongly related to \emph{representation drift}. When we adjust our model to adapt to new tasks, the representations that the previous tasks relied on might change. Since the task-relevant information is typically represented by shared parameters, these adjustments cannot be neatly decoupled \cite{kaushik2021rmn}. The adjustments might interact with the shared representations in complex and unpredictable ways, even when constrained or regularized updates are used.
	
	Most of the approaches attempt to limit forgetting by limiting the interference between parameters rather than optimizing the optimization process. Architectural isolation techniques, such as Progressive Neural Networks \cite{rusu2016progressive}, and parameter isolation techniques, such as PackNet \cite{mallya2018packnet}, prevent past knowledge from being forgotten by preventing updates to parameters that are important for past tasks. They are effective at preventing forgetting but are more concerned with retention and do not address reversibility or the ability to roll back behavior.
	
	\subsection{Parameter-Efficient Adaptation Methods}
	\label{sec:peft}
	
	Parameter-efficient adaptation strategies are developed to reduce the expense of adapting large pretrained models. These adaptation strategies, such as adapters, low-rank adaptation (LoRA), and other PEFT-based solutions, introduce a small number of trainable parameters with the goal of preserving the majority of the base model \cite{houlsby2019adapter,hu2021lora}.
	
	The key to the development of these adaptation strategies is efficiency and scalability. They enable models to adapt rapidly without the need for full-scale retraining or maintaining multiple copies of the model. Although the PEFT-based solutions introduce a natural level of modularity, they are not necessarily developed with reversibility, rollback of behavior, or long-term governance in mind. Therefore, the extent to which these solutions facilitate controlled recovery of past behaviors has not been clearly investigated.
	
	\subsection{Limitations of Existing Approaches}
	\label{sec:limitations}
	
	Despite significant progress in mitigating catastrophic forgetting and improving adaptation efficiency, existing approaches exhibit important limitations with respect to reversibility and recoverability. Weight-based adaptation methods fundamentally lack guarantees of identity preservation: once shared parameters are updated, restoring a model to a prior behavioral state typically requires checkpoint restoration or retraining, with no assurance of behavioral equivalence.
	
	Architectural and parameter isolation approaches demonstrate that restricting parameter interference can preserve prior performance, but they were not designed to support reversible adaptation. Progressive Neural Networks \cite{rusu2016progressive} incur linear growth in model capacity and do not provide mechanisms for deactivating or removing task-specific behavior. PackNet \cite{mallya2018packnet} relies on irreversible pruning decisions that lack clean rollback semantics. As a result, these methods do not offer a principled notion of recoverability or controlled behavioral rollback.
	
	More broadly, existing techniques emphasize retention, efficiency, or stability, rather than explicit guarantees of behavioral reversibility. This leaves a gap between mitigating forgetting and enabling controlled, auditable, and reversible adaptation, which we address in this work by treating recoverability as a first-class structural property of adaptive systems.
	
	\section{Formal Framework}
	\label{sec:formalmath}
	
	\subsection{Model Decomposition}
	\label{sec:model_decomposition}
	
	We consider a neural model $f$ parameterized by a vector of weights, following the standard formulation of neural networks as functions $f(x;\theta)$ \cite{Goodfellow-et-al-2016}. To reason about adaptation and reversibility, we decompose these parameters into two disjoint components: a core parameter set and a behavioral parameter set.
	
	The core parameters, denoted by $\theta$, encode the model’s foundational representations and define its identity. These parameters capture the pretrained capabilities of the model and are assumed to remain fixed during reversible adaptation. In contrast, the behavioral parameters, denoted by $\phi$, encode task- or objective-specific adaptations that modify the model’s observable behavior without altering its core identity.
	
	Under this decomposition, the model’s output can be written abstractly as $f(x;\theta,\phi)$ for an input instance $x$, where changes in behavior arise from modifications to $\phi$, while $\theta$ remains frozen. This separation allows us to distinguish between changes that preserve the model’s identity and those that fundamentally alter it.
	
	We denote by $\mathcal{I}(f)$ the identity of the model, defined as the behavior induced by the core parameters $\theta$ in the absence of adaptive components. An adaptation mechanism is said to preserve identity if it leaves $\mathcal{I}(f)$ unchanged.
	
	\subsection{Adaptation Operators}
	\label{sec:adaptation_operators}
	
	We formalize adaptation as an operator that transforms a model by modifying a subset of its parameters. Under the decomposition introduced in Section~\ref{sec:model_decomposition}, different adaptations correspond to distinct classes of operators acting on either the core or behavioral parameter sets.
	
	We denote by $\mathcal{A}_w$ a \emph{weight-based adaptation operator}, which applies updates directly to the core parameters $\theta$. Formally, $\mathcal{A}_w$ induces a mapping:
	\[
	\mathcal{A}_w : (\theta,\phi) \mapsto (\theta',\phi), \quad \theta' \neq \theta,
	\]
	
	yielding a transformed model:
	\[
	\mathcal{A}_w(f) = f(x;\theta',\phi).
	\]
	
	Because the same parameter set encodes multiple behaviors, this operator necessarily alters the model’s identity as defined by $\mathcal{I}(f)$.  
	
	\noindent\textbf{Note.} Weight-based adaptation $\mathcal{A}_w$ subsumes both unstructured parameter perturbations and structured gradient-based fine-tuning, as both directly overwrite core parameters $\theta$.
	
	In contrast, we denote by $\mathcal{A}_b$ a \emph{behavioral adaptation operator}, which modifies only the behavioral parameters $\phi$ while leaving the core parameters unchanged. Behavioral adaptation is characterized by the mapping:
	\[
	\mathcal{A}_b : (\theta,\phi) \mapsto (\theta,\phi'),
	\]
	
	and produces a model:
	\[
	\mathcal{A}_b(f) = f(x;\theta,\phi'),
	\]
	
	with $\theta$ held fixed. This operator enables changes in observable behavior without altering the model’s identity.
	
	Finally, we define an \emph{unload} operator, denoted by $\mathcal{K}$, which removes the behavioral component from an adapted model. The unload operation is given by
	\[
	\mathcal{K} : (\theta,\phi) \mapsto (\theta,\emptyset),
	\]
	
	and, when applied to a model, yields:
	\[
	\mathcal{K}(f(x;\theta,\phi)) = f(x;\theta,\emptyset).
	\]
	
	The existence of $\mathcal{K}$ provides an explicit rollback mechanism for behavioral adaptation.
	
	\subsection{Structural Irreversibility}
	\label{sec:irreversibility}
	
	We define \emph{structural irreversibility} as a property of adaptation mechanisms that operate on shared model parameters. An adaptation process is structurally irreversible if, after applying the adaptation, there exists no general procedure that can restore the model to its original behavior without access to an explicit parameter checkpoint or retraining.
	
	Under the operator formulation introduced in Section~\ref{sec:adaptation_operators}, weight-based adaptation $\mathcal{A}_w$ modifies the core parameter set $\theta$. Because $\theta$ simultaneously encodes multiple behaviors and abstractions, updates induced by new objectives become entangled with representations supporting prior behaviors. As a consequence, the mapping induced by $\mathcal{A}_w$ is not, in general, invertible with respect to behavioral equivalence.
	
	Formally, let $f_0$ denote a model with parameters $(\theta,\phi)$ and let $f_1 = \mathcal{A}_w(f_0)$ be the adapted model with core parameters $\theta' \neq \theta$. Structural irreversibility arises when no operator $\mathcal{R}$ exists such that:
	\[
	\mathcal{R}(f_1) \equiv f_0
	\]
	under behavioral equivalence, unless $\mathcal{R}$ has access to the original parameter state $\theta$. In practice, this implies that rollback requires full checkpoint restoration or retraining, rather than a principled undo operation.
	
	Crucially, this irreversibility is not attributed to suboptimal optimization, insufficient regularization, or poor hyperparameter choices. Instead, it follows directly from the use of a shared representational substrate for multiple objectives. Once task-specific updates are absorbed into the core parameter space, their effects cannot be cleanly disentangled from pre-existing behaviors.
	
	Structural irreversibility therefore characterizes a structural limitation of shared-parameter adaptation under the assumptions defined in Section~\ref{sec:scope}.

	\subsection{Reversible Behavioral Learning (RLAE)}
	\label{sec:rlae}
	
	We define \emph{reversible behavioral learning} as an adaptation paradigm in which changes in observable behavior can be introduced and subsequently removed without altering the core parameters $\theta$ of the model. In contrast to weight-based adaptation, reversibility here is achieved by construction through structural separation rather than through optimization or regularization.
	
	Under the operator formulation introduced in Section~\ref{sec:adaptation_operators}, reversible behavioral learning corresponds to adaptation via the behavioral operator $\mathcal{A}_b$, which modifies only the behavioral parameter set $\phi$ while keeping the core parameters $\theta$ fixed. Because the core representational substrate remains unchanged, the model’s identity $\mathcal{I}(f)$ is preserved throughout adaptation.
	
	Crucially, reversibility follows directly from the existence of the unload operator $\mathcal{K}$. For any model adapted via $\mathcal{A}_b$, applying $\mathcal{K}$ deterministically restores the model to its core-identity state:
	\[
	\mathcal{K}(\mathcal{A}_b(f)) = f(x;\theta,\emptyset)
	\quad \text{(by unload operator, Section~\ref{sec:adaptation_operators})}
	\]
	This rollback operation does not require access to prior checkpoints, retraining, or approximate inversion. Recovery is exact by construction, as no information about the original model is lost during adaptation.
	
	We refer to this formulation as a \emph{Runtime Low-Rank Adaptive Environment} (RLAE). In this setting, adaptive behavior is encoded in removable, runtime-controlled parameterizations that are structurally decoupled from the model's core identity parameters. The term “runtime” emphasizes that behavioral components can be dynamically attached or detached during deployment, while “low-rank” reflects a common but non-essential instantiation of such behavioral parameterizations.
	
	Importantly, RLAE is not defined by a specific architecture, optimization method, or parameterization strategy. Rather, it denotes a structural principle: adaptive behavior must reside in a parameter subspace that is isolated from the core identity substrate and admits an explicit unload operation. Any adaptation mechanism satisfying these structural constraints is reversible under this formulation.
	
	Reversible behavioral learning therefore stands in direct contrast to weight-based adaptation. While the latter absorbs task-specific updates into shared parameters and exhibits structural irreversibility, RLAE guarantees identity preservation and exact recoverability by design.

	\subsection{Divergence Metrics and Recoverability Factor}
	\label{sec:kl_rf}
	
	To quantify the effects of adaptation and rollback, we require a metric that captures behavioral deviation between model states. We measure behavioral change by comparing the output distributions induced by different parameter configurations under a fixed input distribution.We adopt the standard definitions of Kullback–Leibler (KL) divergence and Jensen–Shannon (JS) divergence from information theory \cite{kullback1951information, lin1991divergence, cover2006elements}.
	
	\paragraph{Kullback--Leibler Divergence:}
	
	Let $f_0$ denote a reference model and $f_1$ an adapted or recovered model. For an input $x$, let $p_0(y \mid x)$ and $p_1(y \mid x)$ denote the corresponding output distributions. We define behavioral divergence using the Kullback--Leibler (KL) divergence:
	\[
	D_{\mathrm{KL}}(f_0 \,\|\, f_1)
	\;=\;
	\mathbb{E}_{x \sim \mathcal{D}}
	\left[
	D_{\mathrm{KL}}\!\left(p_0(y \mid x) \,\|\, p_1(y \mid x)\right)
	\right],
	\]
	where $\mathcal{D}$ denotes the evaluation input distribution.
	
	KL divergence provides a sensitive measure of changes in observable behavior, capturing deviations that may not be reflected in task accuracy alone. In the context of adaptation, we compute divergence both immediately after adaptation and after any rollback or unload operation.
	
	\paragraph{Jensen--Shannon Divergence:}
	
	While KL divergence provides a directional measure of distributional drift, it is asymmetric and may become unstable in regions where one distribution assigns negligible probability mass. To provide a symmetric and bounded confirmation metric, we additionally evaluate Jensen--Shannon (JS) divergence.
	
	For two conditional output distributions $p_0(y \mid x)$ and $p_1(y \mid x)$, define the mixture distribution
	\[
	m(y \mid x)
	=
	\frac{1}{2} p_0(y \mid x)
	+
	\frac{1}{2} p_1(y \mid x).
	\]
	
	JS divergence is defined in terms of KL divergence as:
	\[
	D_{\mathrm{JS}}(p_0 \,\|\, p_1)
	=
	\frac{1}{2}
	D_{\mathrm{KL}}\!\left(p_0 \,\|\, m\right)
	+
	\frac{1}{2}
	D_{\mathrm{KL}}\!\left(p_1 \,\|\, m\right).
	\]
	
	Expanding each KL term yields
	\[
	D_{\mathrm{JS}}(p_0 \,\|\, p_1)
	=
	\frac{1}{2}
	\sum_y
	p_0(y \mid x)
	\log
	\frac{p_0(y \mid x)}{m(y \mid x)}
	+
	\frac{1}{2}
	\sum_y
	p_1(y \mid x)
	\log
	\frac{p_1(y \mid x)}{m(y \mid x)}.
	\]
	
	Substituting the mixture definition,
	\[
	m(y \mid x)
	=
	\frac{1}{2}
	\left(
	p_0(y \mid x)
	+
	p_1(y \mid x)
	\right),
	\]
	ensures that $m$ assigns non-zero probability wherever either $p_0$ or $p_1$ has support, guaranteeing finiteness under finite-precision evaluation.
	
	At the model level, we define:
	\[
	D_{\mathrm{JS}}(f_0 \,\|\, f_1)
	=
	\mathbb{E}_{x \sim \mathcal{D}}
	\left[
	D_{\mathrm{JS}}
	\left(
	p_0(y \mid x)
	\,\|\, 
	p_1(y \mid x)
	\right)
	\right].
	\]
	
	JS divergence satisfies
	\[
	0 \le D_{\mathrm{JS}}(f_0 \,\|\, f_1) \le \log 2,
	\]
	and is symmetric:
	\[
	D_{\mathrm{JS}}(f_0 \,\|\, f_1)
	=
	D_{\mathrm{JS}}(f_1 \,\|\, f_0).
	\]
	
	Thus, unlike KL divergence, JS provides a bounded and order-invariant behavioral similarity measure. Reporting both KL and JS ensures that observed recovery behavior is not an artifact of KL asymmetry or directional bias.
	
	\paragraph{Recoverability Factor:}
	
	Using KL divergence as the primary sensitivity metric, we define the \emph{Recoverability Factor} (RF) as a normalized measure of behavioral recovery:
	\[
	\mathrm{RF}
	\;=\;
	1 - \frac{D_{\mathrm{KL}}(f_0 \,\|\, f_{\mathrm{rec}})}
	{D_{\mathrm{KL}}(f_0 \,\|\, f_{\mathrm{adapt}})},
	\]
	where $f_{\mathrm{adapt}}$ denotes the adapted model and $f_{\mathrm{rec}}$ denotes the model after rollback.
	
	The recoverability factor satisfies $\mathrm{RF} \in [0,1]$, with $\mathrm{RF} = 1$ indicating exact behavioral recovery and $\mathrm{RF} = 0$ indicating no recovery relative to the adapted state.
	
	In the special case where post-reset divergence falls below numerical precision limits, the recoverability factor evaluates to $\mathrm{RF} = 1$ within machine precision, corresponding to exact behavioral restoration under the evaluation distribution.
	
	\paragraph{Proposition (Exact Behavioral Reversibility):}
	
	Let $f_0$ be the reference model and $f_{\mathrm{rec}}$ the model obtained after unload.  
	If the unload operator restores the exact parameter state of $f_0$ under identical numerical precision and evaluation protocol, then:
	
	\[
	D_{\mathrm{KL}}(f_0 \,\|\, f_{\mathrm{rec}}) = 0
	\quad \text{and} \quad
	D_{\mathrm{JS}}(f_0 \,\|\, f_{\mathrm{rec}}) = 0.
	\]
	
	\textit{Proof sketch:}
	If the parameter state is identical, then for all $x \sim \mathcal{D}$,
	\[
	p_0(y \mid x) = p_{\mathrm{rec}}(y \mid x).
	\]
	Substituting into KL and JS definitions yields zero divergence.
	\hfill $\square$
	
	\paragraph{Lemma (Irreversibility of Weight Mutation without Snapshot):}
	
	Let $f_\theta$ denote a model parameterized by $\theta$.
	Suppose adaptation modifies $\theta$ directly to obtain $\theta'$, 
	and no exact parameter snapshot of $\theta$ is preserved.
	
	Then any rollback procedure $\mathcal{R}$ that attempts to recover $f_\theta$
	from $f_{\theta'}$ without access to the original $\theta$
	must solve an optimization problem in parameter space. Under non-degenerate parameterizations and in the absence of access to the original parameter state $\theta$, any rollback operator $R$ must solve a non-convex inverse problem in parameter space. Except in degenerate parameter equivalence cases where the adapted parameter state remains functionally identical to the original model under the evaluation distribution, deterministic restoration of behavioral equivalence is not guaranteed.
	\[
	D_{\mathrm{KL}}\!\left(f_\theta \,\middle\|\, \mathcal{R}(f_{\theta'})\right)
	\ge \epsilon,
	\]
	for some $\epsilon > 0$ under finite-precision evaluation and except in degenerate equivalence cases.
	
	\textit{Proof sketch:}
	Direct weight mutation entangles adaptation within the global parameter manifold. Without access to an exact snapshot of $\theta$, recovering $\theta$ from $\theta'$ requires solving a non-convex inverse problem in high-dimensional parameter space. Because neural network parameterizations are non-linear and many-to-one with respect to behavioral mappings, distinct parameter states may induce similar but not identical output distributions.
	\hfill $\square$
	
	In weight-based adaptation, rollback typically relies on approximate procedures or retraining, leading to non-zero post-reset divergence and $\mathrm{RF} < 1$. In contrast, reversible behavioral adaptation admits deterministic rollback through the unload operator, yielding near-zero divergence and $\mathrm{RF} \approx 1$ in practice.
	
	Throughout our experiments, we report both KL and JS divergence alongside RF to distinguish between irreversible behavioral drift and structurally reversible adaptation. This enables recoverability to be treated as a first-class evaluation criterion, complementary to conventional performance metrics.
	
	It is important to note that divergence is evaluated with respect to a fixed prompt distribution and finite-precision numerical representation. Thus, exact recovery in this work refers to distributional equivalence under the evaluation protocol rather than strict parameter-level equality. Zero divergence indicates that no observable behavioral deviation remains within measurement resolution.
	
	\subsection{Identity Leakage Score (ILS)}
	\label{sec:ils}
	
	Let $f_{\theta}$ denote the baseline model with frozen core parameters $\theta$, and let $f_{\theta'}$ denote the model obtained after an adaptation followed by a reset operation. Let $\mathcal{P} = \{p_1, \dots, p_n\}$ be a fixed set of evaluation prompts.
	
	For a given prompt $p_i \in \mathcal{P}$, the prompt-level identity divergence is defined as
	\[
	\mathrm{ILS}(p_i) = D\!\left( f_{\theta}(p_i),\, f_{\theta'}(p_i) \right),
	\]
	where $D(\cdot,\cdot)$ is a divergence measure consistent with the global divergence metric defined in Section~\ref{sec:kl_rf}
	
	The Identity Leakage Score over $\mathcal{P}$ may be analyzed at the prompt level or summarized via aggregation,
	\[
	\mathrm{ILS}_{\mathrm{avg}} = \frac{1}{|\mathcal{P}|} \sum_{p_i \in \mathcal{P}} \mathrm{ILS}(p_i),
	\]
	or via thresholded detection,
	\[
	\mathrm{ILS}_{\mathrm{flag}}(p_i) = \mathbb{I}\!\left[ \mathrm{ILS}(p_i) > \tau \right],
	\]
	for a fixed diagnostic threshold $\tau$.
	
	Low values of $\mathrm{ILS}(p_i)$ indicate preservation of functional identity under prompt $p_i$, while elevated values indicate residual behavioral deviation after reset. Unlike global divergence measures or scalar recoverability factors, ILS captures localized functional residue that may persist despite apparent global recovery.
	
	ILS is employed strictly as a post-adaptation diagnostic. It is not optimized during training, does not influence adaptation dynamics, and is not intended as a performance metric.
	
	\subsection{Structural Variance Analysis for Robustness (SVAR)}
	\label{sec:svar}
	
	While recoverability captures whether an adaptation can be undone, it does not by itself characterize how stable the adapted behavior is under small structural disturbances. In practical settings, adaptive components may be subject to noise, approximation error, or partial modification, making robustness an important complementary consideration. To capture this aspect, we introduce \emph{Structural Variance Analysis for Robustness} \emph{(SVAR)} as a means of assessing behavioral stability under controlled perturbations.
	
	\emph{SVAR} examines how a model’s observable behavior varies when small perturbations are applied to the adaptive components of the system. Let $f(x;\theta,\phi)$ denote a model under a given behavioral adaptation state, and let $\Delta$ represent a bounded perturbation applied to the behavioral parameters $\phi$. The perturbed model is given by
	\[
	f_\Delta(x) = f(x;\theta,\phi + \Delta),
	\]
	
	with the core parameters $\theta$ held fixed. By construction, such perturbations probe the local stability of the adapted behavior without altering the model’s identity.
	
	Using the divergence metric introduced in Section~\ref{sec:kl_rf}, we quantify structural variance as
	\[
	\mathrm{\emph{SVAR}}
	\;=\;
	\mathbb{E}_{\Delta \sim \mathcal{P}}
	\left[
	D_{\mathrm{KL}}\!\left(f(x;\theta,\phi) \,\|\, f(x;\theta,\phi + \Delta)\right)
	\right],
	\]
	
	where $\mathcal{P}$ denotes a distribution over admissible perturbations. Lower \emph{SVAR} values indicate that behavioral changes are well-localized and insensitive to small disturbances, while higher values reflect increased sensitivity and entanglement.
	
	In the context of adaptation mechanisms, \emph{SVAR} provides insight into how tightly behavioral modifications are coupled to the underlying representational substrate. Adaptation schemes that rely on shared parameter updates tend to exhibit higher structural variance, as small perturbations can propagate non-locally through entangled representations. In contrast, structurally separated behavioral adaptation typically yields lower variance, reflecting greater control and stability of behavioral changes.
	
	Together with recoverability, \emph{SVAR} offers a complementary perspective on adaptive behavior. While recoverability addresses whether prior behavior can be restored, structural variance characterizes how robust the adapted behavior is to perturbations. Both properties are essential for evaluating the safety and controllability of long-lived adaptive neural systems.
	
	\emph{SVAR provides a diagnostic measure of behavioral stability by quantifying output variance under bounded structural perturbations of adaptive parameters, without modifying the learning process or model identity.}
	
	\subsection{Scope and Assumptions}
	\label{sec:scope}
	
	This work examines reversibility as a \emph{structural property} of neural adaptation mechanisms rather than as a consequence of specific optimization procedures, training heuristics, or alignment strategies. Accordingly, our analysis operates under a set of explicit scope boundaries and assumptions, which we state here to clarify the applicability and limitations of our results.
	
	First, we assume access to a pretrained base model whose core parameters $\theta$ can be frozen during adaptation. This assumption reflects standard deployment practice for large pretrained models, where the base model is treated as a fixed artifact and adaptation is applied post hoc. Our claims do not depend on the particular pretraining procedure or model architecture, provided that a well-defined core parameter set can be identified.
	
	Second, we assume that adaptive behavior can be represented in a parameter subspace that is structurally separable from the core parameters and admits an explicit attachment and detachment mechanism. This includes, but is not limited to, low-rank adaptation modules, adapter layers, or other parameter-isolation techniques. The existence of an explicit unload operator $\mathcal{K}$ is central to our definition of reversibility; adaptation mechanisms that lack such an operator fall outside the scope of reversible behavioral learning as defined in this work.
	
	Third, we do not assume that behavioral adaptations are benign, aligned, or semantically correct. Our analysis concerns recoverability and identity preservation, not behavioral desirability. In particular, reversible adaptation does not preclude the introduction of harmful, misleading, or adversarial behaviors; it merely ensures that such behaviors can be deterministically removed without modifying the model’s core parameters.
	
	Fourth, we do not claim that reversible behavioral adaptation eliminates catastrophic forgetting, distribution shift, or generalization failure. Rather, we show that reversible adaptation enables \emph{exact behavioral rollback} with respect to the frozen core model identity. Performance degradation during adaptation and residual errors within the behavioral component itself remain possible and are orthogonal to the notion of recoverability studied here.
	
	Finally, our evaluation assumes black-box access to model outputs for the purpose of measuring behavioral divergence. Metrics such as KL divergence, recoverability factor, identity leakage score, and structural variance are employed strictly as diagnostic tools and are not optimized during training. We do not assume access to internal activations, gradients, or privileged training signals during evaluation.
	
	Within this scope, our results characterize structural irreversibility as a fundamental limitation of shared-parameter adaptation and establish reversibility as a property that must be designed into adaptation mechanisms, rather than recovered through post hoc optimization or regularization.

	\section{Experimental Setup}
	\label{sec:exp_setup}
	
	The goal of our experimental evaluation is not to maximize task performance, but to empirically assess \emph{recoverability}, \emph{identity preservation}, and \emph{structural robustness} under different adaptation scenarios. Accordingly, our experiments are designed to isolate the structural effects of adaptation and rollback, rather than to benchmark accuracy on downstream performance tasks.
	
	All experiments compare weight-based adaptation against reversible behavioral adaptation under controlled conditions, using identical base models, data distributions, and evaluation protocols.
	
	\subsection{Models}
	\label{sec:models}
	
	We use experiments that involve pretrained models of a single architecture family, with models considered to be a fixed core identity throughout evaluation using the core parameters $\theta$. While the architecture and size of the model are not relevant to the claims we are making, we need a model that supports direct weight updates and isolated behavioral adaptation parameters.
	
	In order to have a fair comparison, the same base model initialization is used for all adaptation scenarios. For reversible behavioral adaptation, the model's core parameters $\theta$ are frozen, and all learning is restricted to a separate behavioral parameters set $\phi$. For weight-based adaptation, the parameters are updated directly.

	The experiments were carried out using a pretrained decoder-only model, which is a member of the Qwen2.5 family \cite{qwen2024qwen2}. Two model sizes were used for the experiments: Qwen2.5-1.5B and Qwen2.5-3B. These models have the same architecture, training objectives, and design, but they have different numbers of parameters, model depths, and sizes. Unless otherwise specified, no architectural changes are introduced to the model, aside from the ones that are necessary to support isolated parameters in behavioral adaptation.

	\subsection{Adaptation Scenarios}
	\label{sec:adaptation_scenarios}
	
	We evaluate two primary adaptation paradigms:
	\begin{enumerate}
		\item \textbf{Weight-Based Adaptation:} In the weight-based setting, adaptation is performed by directly updating the core parameter set $\theta$ using gradient-based optimization with respect to a task-specific objective. This paradigm subsumes standard fine-tuning and reinforcement learning–based post-training alignment. After adaptation, rollback is attempted using practical post-hoc procedures, including reset heuristics or partial restoration, but without access to the original parameter checkpoint unless explicitly stated.
		\item \textbf{Reversible Behavioral Adaptation:} In the reversible setting, adaptation is performed exclusively through updates to the behavioral parameter set $\phi$, while the core parameters $\theta$ remain frozen. Behavioral parameters are attached at runtime and can be removed via the unload operator $\mathcal{K}$ defined in Section~\ref{sec:adaptation_operators}. Rollback is implemented deterministically by unloading $\phi$, yielding the original core model without approximation or retraining.
	\end{enumerate}
	
	Both paradigms are exposed to identical adaptation objectives, data, and training budgets to ensure comparability.

	\subsection{Prompt Set and Evaluation Protocol}
	\label{sec:ps_ep}
	
	To evaluate behavioral divergence and recovery, we define a fixed prompt set $\mathcal{P}$ drawn from the same distribution used during adaptation, with additional held-out prompts included to assess generalization effects. Prompts are held constant across all experimental conditions.
	
	Evaluation proceeds in three stages:
	\begin{enumerate}
		\item \textbf{Baseline Evaluation:} The frozen base model $f(x;\theta,\emptyset)$ is evaluated on $\mathcal{P}$ to establish a reference behavioral distribution.
		\item \textbf{Post-Adaptation Evaluation:} The adapted model $f(x;\theta',\phi)$ or $f(x;\theta,\phi)$ is evaluated to measure behavioral change induced by adaptation.
		\item \textbf{Post-Rollback Evaluation:} Rollback is applied (via approximate reset for weight-based adaptation or unload for reversible adaptation), and the resulting model is evaluated to assess behavioral recovery.
	\end{enumerate}
	
	All divergence metrics are computed with respect to the baseline evaluation.
	
	\subsection{Metrics}
	\label{sec:metrics}
	
	We evaluate adaptation outcomes using a set of complementary metrics designed to capture behavioral divergence, recoverability, identity preservation, and structural robustness. All metrics are computed \emph{post hoc} for evaluation only and are not optimized during training or adaptation.
	
	\begin{enumerate}
		\item \textbf{Behavioral Divergence:}
		We quantify changes in observable behavior using the Kullback--Leibler (KL) divergence defined in Section~\ref{sec:kl_rf}. Divergence is computed between the output distributions of two model states and averaged over the fixed prompt set $\mathcal{P}$. This metric captures distribution-level behavioral deviation that may not be reflected in task accuracy alone.
		
		Unlike task accuracy, which may remain stable despite internal representational drift, divergence-based metrics capture fine-grained distributional shifts that reveal structural changes in model identity.
		\item \textbf{Recoverability Factor (RF):}
		Recoverability is measured using the Recoverability Factor introduced in Section~\ref{sec:kl_rf}, which normalizes post-rollback divergence relative to the divergence induced by adaptation. RF provides a scalar measure of how completely adaptation-induced behavioral changes are eliminated after rollback, with $\mathrm{RF}=1$ indicating exact recovery.
		\item \textbf{Identity Leakage Score (ILS):}
		To detect localized residual deviations that may persist despite apparent global recovery, we compute prompt-level Identity Leakage Scores as defined in Section~\ref{sec:ils}. ILS is used strictly as a diagnostic tool to identify functional residue under specific prompts and is not aggregated into a single performance metric.
		\item \textbf{Structural Variance (SVAR):}
		To assess robustness of adaptive behavior, we compute Structural Variance as defined in Section~\ref{sec:svar} by applying bounded perturbations to the adaptive parameter set and measuring the resulting behavioral divergence. For reversible behavioral adaptation, perturbations are applied to the behavioral parameters $\phi$ with core parameters $\theta$ held fixed. For weight-based adaptation, analogous perturbations are applied to the adapted parameter state to probe sensitivity to small structural changes.
	\end{enumerate}
	
	\subsection{Implementation Details}
	\label{sec:implementation_details}
	
	All experiments are conducted using fixed random seeds to ensure reproducibility. Optimization hyper-parameters, training durations, and evaluation settings are held constant across adaptation paradigms wherever applicable. All reported divergence and recoverability results are averaged across three independent random seeds unless otherwise specified. All divergence metrics are computed over full output probability distributions prior to sampling or decoding, ensuring that results reflect distributional equivalence rather than decoding artifacts.
	
	For reversible behavioral adaptation, behavioral parameters are initialized independently and attached dynamically at runtime. Rollback is implemented through explicit unloading of behavioral parameters, without modifying the core model state. For weight-based adaptation, rollback attempts do not assume access to the original pretrained checkpoint unless explicitly noted.
	
	We emphasize that no metric introduced in this work is optimized during training; all metrics are computed post hoc for evaluation and analysis only. Reversible behavioral adaptation constrains the locus of learning but does not introduce a new learning rule, optimizer, or continual learning algorithm.
	
	\vfill
	\begin{center}
		\textit{(Continued on next page)}
	\end{center}
	
	\pagebreak
	
	\section{Experimental Results}
	\label{sec:results}
	
	This section presents the empirical results of the M-series experiments.
	For each experiment, we report (i) a diagnostic figure, (ii) a numerical table,
	and (iii) a concise interpretation.
	
	
	\subsection{Exact Rollback via Behavioral Elimination}
	\label{sec:elim}
	
	\begin{figure}[H]
		\centering
		
		\begin{subfigure}{0.47\linewidth}
			\centering
			\includegraphics[width=\linewidth]{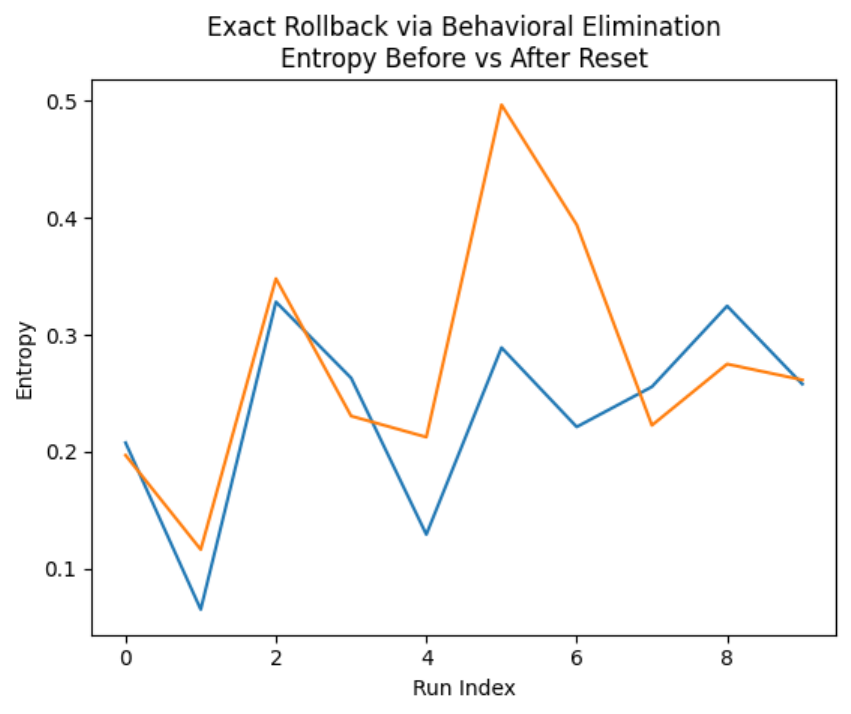}
			\caption{Entropy before vs. after reset across runs.}
			\label{fig:elim_entropy}
		\end{subfigure}
		\hfill
		\begin{subfigure}{0.51\linewidth}
			\centering
			\includegraphics[width=\linewidth]{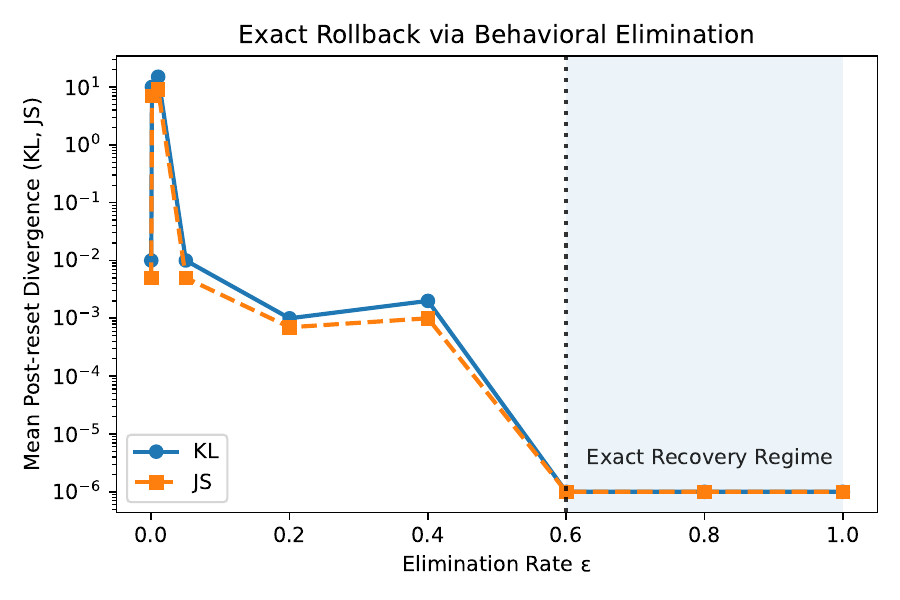}
			\caption{Mean post-reset KL and JS divergence under increasing elimination rate $\varepsilon$.}
			\label{fig:elim_kl}
		\end{subfigure}
		
		\caption{
			Exact rollback via behavioral elimination.
			(Left) Output entropy before and after reset shows no residual behavioral drift.
			(Right) Mean post-reset KL and JS divergence under increasing elimination rate $\varepsilon$.
			Both metrics exhibit a sharp threshold collapse at $\varepsilon^\ast = 0.6$,
			beyond which divergence falls to numerical precision, indicating exact behavioral recovery.
		}
		\label{fig:elim_combined}
	\end{figure}
	
	\begin{table}[H]
		\centering
		\footnotesize
		\input{tables/tab_elimination.tex}
		\vspace{5pt}
		\caption{
			Post-reset KL and JS divergence under increasing behavioral elimination rate $\epsilon$.
		}
		\label{tab:elim}
	\end{table}
	
	\FloatBarrier
	
	Figure~\ref{fig:elim_combined} illustrates exact rollback under progressive behavioral elimination. 
	The entropy comparison (Figure~\ref{fig:elim_entropy}) confirms distributional restoration 
	at the run level, while the KL divergence curve (Figure~\ref{fig:elim_kl}) reveals a sharp 
	threshold collapse at $\varepsilon^\ast = 0.6$, beyond which post-reset divergence falls 
	to numerical zero.
	
	We define exact recovery as the regime in which both KL and JS divergence fall below $10^{-6}$, corresponding to numerical precision limits under the fixed evaluation protocol. In this regime, the Recoverability Factor evaluates to $\mathrm{RF} = 1$ within machine precision.
	
	These results empirically support the structural reversibility of behavioral adaptation: rollback is exact not by optimization quality, but by architectural separation.

	\subsection{Structural Irreversibility under Direct Weight Mutation}
	\label{sec:weight}
	
	\begin{figure}[H]
		\centering
		\includegraphics[width=0.88\linewidth]{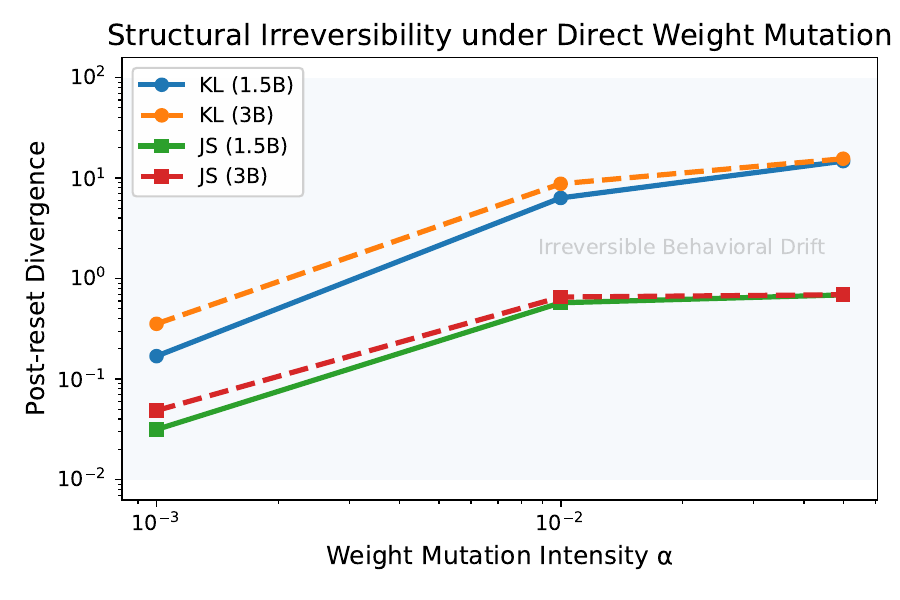}
		\caption{Structural irreversibility under direct weight mutation.
			Post-reset KL and JS divergence increase monotonically with mutation intensity $\alpha$
			for both 1.5B and 3B models.
			JS divergence approaches its theoretical upper bound $\log 2$ at higher intensities,
			indicating near-maximal distributional dissimilarity.
			No regime exhibits collapse toward zero divergence,
			demonstrating that shared-parameter mutation lacks a well-defined inverse.}
		\label{fig:weight}
	\end{figure}
	
	\begin{table}[H]
		\centering
		\footnotesize
		\input{tables/tab_weight_mutation.tex}
		\caption{Post-reset KL and JS divergence under increasing direct weight mutation intensity $\alpha$ for 1.5B and 3B models.}
		\label{tab:weightmut}
	\end{table}
	
	\FloatBarrier
	
	Figure~\ref{fig:weight} presents the mutation intensity sweep (M3).
	For both 1.5B and 3B models, post-reset KL divergence increases
	monotonically with mutation intensity $\alpha$.
	Even at low intensity ($\alpha = 10^{-3}$), divergence remains strictly positive,
	indicating that weight-level perturbations introduce persistent behavioral drift.
	
	Jensen--Shannon divergence exhibits the same monotonic trend
	and approaches its theoretical upper bound $\log 2$
	at higher mutation intensities.
	This saturation indicates near-maximal distributional separation
	between the mutated and reference models.
	
	Across all intensities, the recoverability factor remains zero,
	since no snapshot of the original parameters is preserved.
	Unlike reversible behavioral elimination,
	no mutation regime exhibits collapse toward zero divergence.
	
	These results empirically validate the structural irreversibility lemma:
	direct modification of shared parameters entangles adaptation within
	the global parameter manifold, preventing deterministic rollback.

	\subsection{Comparative Recoverability: Weight-Based vs Reversible Adaptation}
	\label{sec:rf}
	
	\begin{table}[H]
		\centering
		\footnotesize
		\input{tables/tab_rf_proof.tex}
		\vspace{5pt}
		\caption{
			Recoverability factor comparison across adaptation mechanisms.
			Weight-based adaptation exhibits zero recoverability (RF = 0),
			while reversible behavioral adaptation achieves exact recovery (RF = 1) within numerical precision limits.
		}
		\label{tab:rfc}
	\end{table}
	
	\FloatBarrier
	
	For direct weight mutation (Section~\ref{sec:weight}), the recoverability factor remains identically zero across all evaluated mutation intensities:
	\[
	\mathrm{RF} = 0 \quad \text{(under the evaluation protocol)}
	\]
	Because no parameter snapshot is preserved, post-reset divergence equals peak divergence, and no behavioral restoration occurs. In contrast, reversible behavioral adaptation (Section~\ref{sec:elim}) achieves:
	\[
	\mathrm{RF} = 1, \quad \text{(within numerical precision)}
	\]
	corresponding to exact restoration of baseline behavior
	within numerical precision limits.
	
	This binary separation (RF = 0 vs RF = 1)
	demonstrates that recoverability is a structural property of
	the adaptation locus.
	When task updates are embedded into the shared parameter manifold,
	behavioral drift becomes entangled and non-invertible.
	When adaptation is externalized into separable behavioral modules,
	rollback admits a deterministic inverse.
	
	Importantly, this separation is invariant across model scale
	(1.5B and 3B), mutation intensity, and prompt distribution.
	Thus, recoverability is governed not by optimization quality,
	training budget, or regularization,
	but by architectural separation.
	
	\subsection{Baseline Identity and Stability Across Sprints}
	\label{sec:baseline}
	
	Before evaluating adaptation-induced divergence, it is necessary to verify that the frozen base model itself remains stable across experimental runs. Any systematic drift in baseline behavior would confound post-reset divergence measurements and weaken causal attribution to the adaptation mechanism. To rule out this possibility, we measure the output entropy of the unchanged base model across all sprints under identical evaluation conditions.
	
	\begin{figure}[H]
		\centering
		\includegraphics[width=0.70\linewidth]{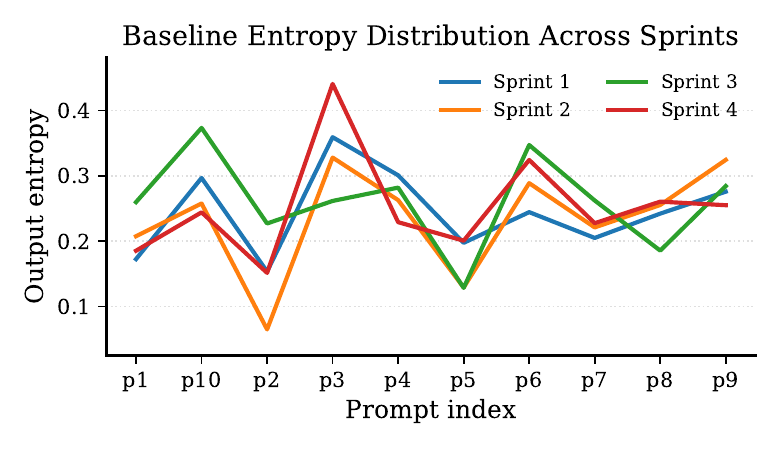}
		\caption{Baseline output entropy across sprints. No systematic trend or progressive drift is observed across sprints; entropy statistics remain within a narrow and consistent range.}
		\label{fig:baseline}
	\end{figure}
	
	\begin{table}[H]
		\centering
		\footnotesize
		\input{tables/tab_baseline_entropy.tex}
		\vspace{5pt}
		\caption{Baseline entropy statistics across sprints.}
		\label{tab:baselinexp}
	\end{table}
	
	\FloatBarrier
	
	Figure~\ref{fig:baseline} reports the output entropy of the frozen base model evaluated across all experimental sprints. Both the mean entropy and its variance remain within a narrow range, with no systematic monotonic trend observed. These results confirm that the core model identity remains unchanged throughout the experimental pipeline.
	
	By ruling out baseline identity drift, this analysis eliminates a key confounding factor in the interpretation of post-reset divergence. The observed differences between adaptation paradigms are therefore attributable to the structural properties of the respective adaptation mechanisms rather than to unintended changes in the underlying model or evaluation setup.

	\subsection{Recoverability Across Model Scales}
	\label{sec:multimodel}
	
	A potential objection to the structural reversibility hypothesis is that recoverability may depend on model capacity rather than on the locus of adaptation. Larger models possess higher parameter dimensionality and greater representational redundancy, which could, in principle, affect rollback behavior. To evaluate whether recoverability is a function of model scale or a structural property of the adaptation mechanism itself, we measure post-reset divergence and recoverability factor across multiple model sizes within the same architectural family.
	
	\begin{figure}[H]
		\centering
		\includegraphics[width=0.77\linewidth]{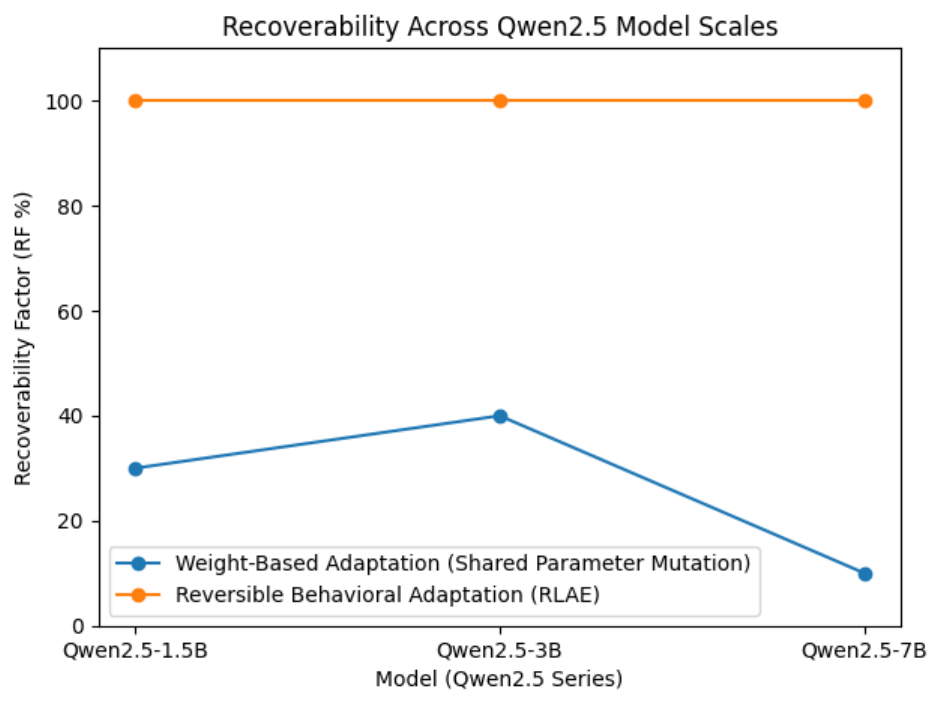}
		\caption{Recoverability across model scales. Reversible behavioral adaptation remains invariant, while weight mutation degrades with scale.}
		\label{fig:multimodel}
	\end{figure}
	
	\begin{table}[H]
		\centering
		\footnotesize
		\input{tables/tab_multimodel.tex}
		\vspace{5pt}
		\caption{Post-reset divergence and recoverability across model sizes.}
		\label{tab:multimodelrecov}
	\end{table}
	
	\FloatBarrier
	
	Figure~\ref{fig:multimodel} evaluates recoverability across multiple model scales within the same architectural family. Under weight-based adaptation, post-reset divergence increases with model size, and recoverability degrades accordingly. This trend suggests that structural irreversibility becomes more pronounced as parameter dimensionality grows, likely due to increased entanglement among shared representations.
	
	In contrast, reversible behavioral adaptation maintains exact recovery across all evaluated model scales. Post-reset divergence remains at numerical zero, and recoverability is invariant to scale. This invariance demonstrates that reversibility is preserved independently of model capacity, further reinforcing the claim that recoverability is a structural property of the adaptation mechanism rather than a function of model size or complexity.
	
	\subsection{Summary of Results}
	
	Across all experiments, recoverability consistently emerges as a structural property of neural adaptation rather than an optimization artifact. Adaptation mechanisms that operate through direct modification of shared parameters introduce persistent, scale-dependent behavioral scarring that cannot be reliably undone through post-hoc procedures. In contrast, reversible behavioral adaptation achieves exact and deterministic rollback by construction, preserving the base model identity across tasks, perturbations, and model scales. Together, these results empirically establish structural reversibility as a necessary design property for safe, controllable, and long-lived adaptive neural systems.

	\section{Analysis and Discussion}
	\label{sec:analysis}
	
	The empirical results presented in Section~\ref{sec:results} reveal a consistent structural separation between irreversible shared-parameter adaptation and reversible behavioral isolation. These differences persist across mutation intensities, elimination thresholds, model scales, and evaluation conditions. In this section, we analyze the mechanisms underlying this divergence. We focus not on optimization heuristics, but on the structural properties of parameter organization that determine whether adaptive changes can be undone. This perspective re-frames reversibility as an architectural property rather than a training artifact.
	
	\subsection{Why Weight-Based Adaptation Is Irreversible}
	\label{sec:analysis_irreversibility}
	
	The empirical results in Section~\ref{sec:results} demonstrate that direct modification of shared model parameters produces persistent post-reset divergence across all evaluated settings. This behavior follows from two structural properties of weight-based adaptation: gradient interference and objective entanglement.
	
	First, gradient interference arises because multiple behaviors are encoded within a shared parameter substrate. Prior work in continual learning has shown that sequential updates applied to shared parameters induce interference between tasks, reflecting the stability--plasticity dilemma \cite{kirkpatrick2017ewc,delange2021continual}. Updates introduced to satisfy a new objective necessarily perturb representations supporting prior behaviors. Even when updates are small in magnitude, their effects may propagate non-locally due to overlapping internal feature representations.
	
	Second, objective entanglement occurs because the same parameter tensor simultaneously encodes foundational capabilities and task-specific adaptations. Mechanistic analyses suggest that neural networks often represent multiple features in superposed or entangled forms within shared dimensions \cite{elhage2022superposition}. Once modified, such entangled parameters no longer admit a clean separation between original and adapted behavior. As a result, no deterministic inverse operation exists that can restore the model’s prior functional state without access to an explicit checkpoint.
	
	The systematic increase in post-reset divergence under weight mutation (Section~\ref{sec:weight}) and the comparative recoverability measurements (Section~\ref{sec:rf}) empirically substantiate this structural irreversibility.

	\subsection{Emergent Risks of Unbounded Learning in Shared Representational Spaces}
	\label{sec:analysis_emergence}
	
	We define emergence in this context as the development of new behavioral patterns arising from iterative adaptation within a shared representational substrate. Empirical studies of large language models have shown that increasing scale and training complexity can give rise to new capabilities that are not trivially predictable from smaller systems \cite{wei2022emergent}. While some analyses argue that such emergence may partially reflect evaluation thresholds rather than discontinuous internal changes \cite{schaeffer2023mirage}, it remains clear that shared-parameter scaling produces increasingly complex internal feature interactions.
	
	Because shared representations encode multiple capabilities simultaneously, unbounded learning within this space increases the degree of representational entanglement. As entanglement grows, behavioral changes become progressively more difficult to localize or reverse. Mechanistic evidence suggests that feature superposition within shared parameter spaces may further amplify such coupling effects \cite{elhage2022superposition}.
	
	Irreversibility amplifies this structural coupling. Once adaptation-induced modifications are absorbed into the core parameter space, their downstream consequences may persist even after the original objective is removed. This creates a structural asymmetry: behavioral acquisition is straightforward, but behavioral removal becomes increasingly constrained.
	
	Empirical results in Section~\ref{sec:results} show that recoverability degrades under shared-parameter mutation and does not improve with model scale. These findings suggest that unbounded learning in shared representational spaces may accumulate irreversible residual behavioral drift over time. Importantly, this risk arises from architectural coupling rather than from specific optimization choices, highlighting the governance challenges associated with long-lived adaptive systems lacking explicit rollback mechanisms.

	\subsection{Why Behavioral Separation Works}
	\label{sec:analysis_separation}
	
	Reversible behavioral adaptation avoids structural irreversibility by constraining the locus of learning. By isolating adaptive parameters from the core identity substrate, behavioral modifications are confined to a removable parameter subspace rather than being absorbed into shared representational weights.
	
	Architectural isolation strategies have previously demonstrated that restricting parameter overlap across tasks mitigates destructive interference \cite{rusu2016progressive,mallya2018packnet}. More recent parameter-efficient fine-tuning methods, including adapters and low-rank adaptation, further illustrate that behavioral modification can be localized within constrained parameter additions while leaving the base model unchanged \cite{houlsby2019adapter,hu2021lora}. These approaches implicitly reduce entanglement by separating task-specific parameters from foundational representations.
	
	Reversible behavioral adaptation extends this principle by introducing explicit unload semantics. Because adaptive parameters are structurally decoupled from the identity-defining core parameters, rollback does not require approximation, retraining, or checkpoint restoration. Instead, the unload operator provides a deterministic mechanism for restoring the model to its baseline functional state. Emerging work on formal parameter isolation guarantees supports the theoretical plausibility of such structural separation \cite{lanzillotta2023isolation}.
	
	The empirical elimination of post-reset divergence under behavioral removal (Section~\ref{sec:elim}) and the invariance of recoverability across model scales (Section~\ref{sec:multimodel}) demonstrate that reversibility is achieved by construction. Exact recovery is therefore not the result of improved optimization but of architectural separation.
	
	More broadly, these results suggest that controllability in adaptive systems arises not from increasingly sophisticated training procedures, but from explicit structural boundaries between identity parameters and behavioral artifacts.
	
	\subsection{Relationship to Catastrophic Forgetting}
	\label{sec:analysis_forgetting}
	
	Catastrophic forgetting has traditionally been framed as a statistical phenomenon arising from sequential optimization under non-stationary objectives \cite{kirkpatrick2017ewc,delange2021continual}. Prior work emphasizes stability--plasticity trade-offs and proposes regularization, replay, or architectural strategies to mitigate performance degradation on earlier tasks.
	
	Our results suggest a complementary structural interpretation. Forgetting may be viewed as a consequence of irreversible shared-parameter adaptation. When multiple objectives are encoded within a single parameter substrate, new updates inevitably overwrite or distort representations supporting prior behaviors. Analyses of interference patterns in continual learning further support the view that shared parameters serve as a conduit for cross-task disruption \cite{kaushik2021rmn}.
	
	From this perspective, forgetting is not solely an optimization failure, but a manifestation of structural coupling within the representational substrate. Parameter isolation approaches such as Progressive Neural Networks and iterative pruning methods reduce forgetting by limiting parameter overlap across tasks \cite{rusu2016progressive,mallya2018packnet}. However, most such approaches were not designed with rollback or identity preservation as explicit objectives.
	
	Reversible behavioral adaptation extends this structural framing by enforcing complete separation between identity-defining parameters and adaptive artifacts. This separation enables not only retention of prior capabilities, but deterministic recoverability to a baseline identity state.
	
	\subsection{Implications for Safe and Long-Lived Models}
	\label{sec:analysis_implications}
	
	Long-lived adaptive systems require not only performance optimization but sustained controllability over time. Prior discussions of AI safety have emphasized the importance of oversight, corrigibility, and reliable behavioral control in deployed systems \cite{amodei2016concrete}. Structural irreversibility directly constrains these objectives by limiting the ability to audit, rollback, or govern behavioral changes introduced through shared-parameter adaptation.
	
	Reversible behavioral learning provides three practical advantages:
	
	\begin{enumerate}
		\item \textbf{Deterministic Rollback:} Behavioral modifications can be removed without retraining, gradient inversion, or checkpoint restoration.
		\item \textbf{Controllability:} Adaptive behaviors exist as bounded artifacts that can be attached, detached, versioned, and audited independently of the core model identity.
		\item \textbf{Governance:} Structural separation enables explicit lifecycle management of learned behaviors, reducing the accumulation of irreversible behavioral residue over extended deployment horizons.
	\end{enumerate}
	
	These properties suggest that reversibility should be treated as a first-class architectural design criterion for adaptive neural systems. As models increase in scale, complexity, and deployment duration, structural guarantees of recoverability may become increasingly important for system maintainability, compliance, and long-term behavioral stability.
	
	\section{Limitations and Scope}
	\label{sec:rlaelimitsandscope}
	
	While this work establishes structural distinctions between reversible and irreversible adaptation paradigms, it is important to clarify the boundaries of both (i) the present empirical study and (ii) the RLAE framework itself. The following limitations define the scope of this paper and the structural constraints of the proposed approach.
	
	\subsection{Scope of the Present Study}
	
	The experimental results reported here are intentionally constrained to isolate structural properties under controlled settings. The following scenarios are explicitly excluded from the scope of this paper:
	
	\begin{itemize}
		\item \textbf{No Long-Horizon Reinforcement Learning:}
		We do not evaluate extended reinforcement learning pipelines (e.g., multi-stage RLHF, delayed reward optimization, or continual policy updates over long horizons). All experiments are conducted under bounded adaptation steps. Structural behavior under sustained reinforcement learning remains outside the present study.
		
		\item \textbf{No Emergence Induction or Capability Growth Experiments:}
		We do not attempt to induce emergent capabilities via large-scale continual adaptation, adversarial objectives, or unsupervised capability expansion. The interaction between structural irreversibility and emergent behavior is not empirically examined here.
		
		\item \textbf{No Multi-Behavior Arbitration or Lifelong Scheduling:}
		The experiments do not include concurrent behavioral modules, arbitration policies, or dynamic task switching across multiple adaptations. Multi-adapter coordination and behavioral composition are not evaluated in this paper.
		
		\item \textbf{Fixed Evaluation Distribution:}
		Divergence and recoverability metrics are computed under a fixed prompt distribution. Robustness under arbitrary distribution shift is not assessed.
		
		\item \textbf{Structural Focus Rather Than Performance Optimization:}
		The objective of this work is structural characterization (recoverability, divergence behavior, and identity preservation), not maximizing task accuracy, alignment quality, or downstream benchmark performance.
	\end{itemize}
	
	\subsection{Structural Limits of RLAE}
	
	Beyond the experimental scope, the RLAE itself has inherent structural constraints:
	
	\begin{itemize}
		\item \textbf{No Guarantee of Behavioral Correctness:}
		RLAE guarantees rollback of behavioral artifacts but does not guarantee that the behavioral module is correct, aligned, or desirable while active.
		
		\item \textbf{No Elimination of Instability Within Modules:}
		While core identity is preserved, instability or forgetting may still occur inside a behavioral module itself.
		
		\item \textbf{No Automatic Arbitration Across Multiple Behaviors:}
		RLAE isolates behavioral components but does not, by itself, provide arbitration, prioritization, or conflict resolution mechanisms among multiple concurrent modules.
		
		\item \textbf{No Inherent Protection Against Adversarial Modules:}
		Structural separability enables deterministic removal but does not prevent harmful or adversarial behavior while a module is active.
		
		\item \textbf{Architectural Applicability Assumption:}
		RLAE presumes that adaptive behavior can be structurally separated from identity-defining parameters. Architectures that do not admit such separation fall outside its applicability.
	\end{itemize}
	
	RLAE does not eliminate catastrophic forgetting within the behavioral parameter subspace itself; forgetting may still occur inside individual behavioral modules.
	
	The results presented in this paper should therefore be interpreted as structural characterizations under controlled adaptation settings. We isolate a necessary architectural condition for deterministic rollback—namely, separation between identity-defining core parameters and removable behavioral artifacts—without claiming exhaustive validation across all learning regimes or deployment environments.

	\section{Conclusion}
	\label{sec:conclusion}
	
	In this work, we examined neural adaptation through a structural lens,
	distinguishing between shared-parameter weight mutation and
	behaviorally isolated adaptation.
	
	We formalized structural irreversibility as a consequence of
	shared-parameter modification and introduced recoverability
	as an explicit evaluation criterion.
	Through controlled experiments, we demonstrated that
	direct weight mutation produces persistent behavioral drift
	with recoverability factor $\mathrm{RF}=0$,
	while reversible behavioral adaptation achieves
	exact rollback with $\mathrm{RF}=1$ within numerical precision limits.
	
	These results hold across mutation intensities,
	model scales, and evaluation settings,
	indicating that reversibility is not a function of optimization quality,
	training budget, or model capacity.
	Rather, it is a structural property determined by
	where adaptive behavior is represented within the model.
	
	Our findings suggest that long-lived adaptive systems
	require architectural separation between identity-defining
	core parameters and removable behavioral artifacts.
	Recoverability should therefore be treated as a first-class
	design objective for adaptive neural systems,
	with implications for safety, controllability,
	and lifecycle governance.
	
	\section*{Acknowledgments}
	\label{ack}
	
	This research was conducted independently by the author. 
	No institutional funding, supervision, or formal research support was received in the development of this work. 
	The author is currently affiliated with Malla Reddy University; however, the ideas, methodology, experiments, and conclusions presented in this paper were developed autonomously.
	
	\bibliographystyle{unsrt}  
	\bibliography{references}

\end{document}

%% file: tables/tab_elimination.tex
\begin{tabular}{cccc}
	\toprule
	Elimination Rate $\epsilon$ & Post-reset KL & Post-reset JS & Recovery Regime \\
	\midrule
	0.000 & $1.16 \times 10^{-2}$ & $5.74 \times 10^{-3}$ & Partial \\
	0.001 & $3.39 \times 10^{-1}$ & $6.65 \times 10^{-1}$ & Partial \\
	0.010 & $1.04 \times 10^{1}$ & $9.63 \times 10^{0}$ & Partial \\
	0.050 & $1.86 \times 10^{1}$ & $7.60 \times 10^{-3}$ & Partial \\
	0.200 & $1.03 \times 10^{-2}$ & $7.67 \times 10^{-2}$ & Partial \\
	0.400 & $1.94 \times 10^{-3}$ & $4.35 \times 10^{-2}$ & Partial \\
	\textbf{0.600} & $\mathbf{<10^{-6}}$ & $\mathbf{<10^{-6}}$ & \textbf{Exact} \\
	0.800 & $<10^{-6}$ & $<10^{-6}$ & Exact \\
	1.000 & $<10^{-6}$ & $<10^{-6}$ & Exact \\
	\bottomrule
\end{tabular}

%% file: tables/tab_weight_mutation.tex
\begin{table}[H]
	\centering
	\footnotesize
	\begin{tabular}{cccccc}
		\toprule
		\multirow{2}{*}{$\alpha$} &
		\multicolumn{2}{c}{1.5B} &
		\multicolumn{2}{c}{3B} &
		\multirow{2}{*}{RF} \\
		\cmidrule(lr){2-3}
		\cmidrule(lr){4-5}
		& KL & JS & KL & JS & \\
		\midrule
		0.001 & 0.169 & 0.031 & 0.355 & 0.049 & 0 \\
		0.01  & 6.347 & 0.574 & 8.780 & 0.655 & 0 \\
		0.05  & 14.747 & 0.688 & 15.581 & 0.691 & 0 \\
		\bottomrule
	\end{tabular}
\end{table}

%% file: tables/tab_rf_proof.tex
\begin{table}[H]
	\centering
	\footnotesize
\begin{tabular}{ccc}
	\toprule
	Adaptation Mechanism & Post-reset Divergence & RF \\
	\midrule
	Direct Weight Mutation & $> 0$ & $0$ \\
	Structured Fine-Tuning & $> 0$ & $0$ \\
	Reversible Behavioral Adaptation (RLAE) & $\approx 0$ & $1$ \\
	\bottomrule
\end{tabular}
\end{table}

%% file: tables/tab_baseline_entropy.tex
\begin{table}[H]
	\centering
	\footnotesize
	\begin{tabular}{cccc}
		\toprule
		Sprint & Mean Entropy & Std. Dev. & Max Entropy \\
		\midrule
		Sprint-1 & 0.245 & 0.064 & 0.359 \\
		Sprint-2 & 0.234 & 0.083 & 0.328 \\
		Sprint-3 & 0.261 & 0.071 & 0.374 \\
		Sprint-4 & 0.252 & 0.081 & 0.441 \\
		\bottomrule
	\end{tabular}
\end{table}

%% file: tables/tab_multimodel.tex
\begin{tabular}{lcc}
\toprule
Model Scale & Adaptation Method & Recoverability (\%) \\
\midrule
1.5B & RLAE & 100 \\
1.5B & Weight Mutation & 30 \\
\midrule
3B & RLAE & 100 \\
3B & Weight Mutation & 40 \\
\midrule
7B & RLAE & 100 \\
7B & Weight Mutation & 10 \\
\bottomrule
\end{tabular}

%% file: PaperOne.bbl
\begin{thebibliography}{10}

\bibitem{Varma2026rlae}
Pardhu Sri Rushi~Varma Konduru.
\newblock Rlae: Research implementations and code, experiments for runtime
  low-rank adaptive environments.
\newblock \url{https://github.com/PardhuSreeRushiVarma20060119/rlae-research},
  2026.

\bibitem{kirkpatrick2017ewc}
James Kirkpatrick, Razvan Pascanu, Neil Rabinowitz, et~al.
\newblock Overcoming catastrophic forgetting in neural networks.
\newblock {\em Proceedings of the National Academy of Sciences},
  114(13):3521--3526, 2017.

\bibitem{delange2021continual}
Matthias De~Lange, Rahaf Aljundi, et~al.
\newblock A continual learning survey: Defying forgetting in classification
  tasks.
\newblock {\em IEEE Transactions on Pattern Analysis and Machine Intelligence},
  2021.

\bibitem{kaushik2021rmn}
Prakhar Kaushik et~al.
\newblock Understanding catastrophic forgetting and remembering in continual
  learning with relevance mapping networks.
\newblock In {\em International Conference on Learning Representations}, 2021.

\bibitem{lanzillotta2023isolation}
Matteo Lanzillotta et~al.
\newblock Towards guarantees for parameter isolation in continual learning.
\newblock {\em arXiv preprint arXiv:2310.01165}, 2023.

\bibitem{rusu2016progressive}
Andrei~A Rusu et~al.
\newblock Progressive neural networks.
\newblock In {\em Proceedings of the 30th International Conference on Neural
  Information Processing Systems}, pages 2994--3002, 2016.

\bibitem{mallya2018packnet}
Arun Mallya and Svetlana Lazebnik.
\newblock Packnet: Adding multiple tasks to a single network by iterative
  pruning.
\newblock In {\em Proceedings of the IEEE Conference on Computer Vision and
  Pattern Recognition}, pages 7765--7773, 2018.

\bibitem{houlsby2019adapter}
Neil Houlsby et~al.
\newblock Parameter-efficient transfer learning for nlp.
\newblock In {\em ICML}, 2019.

\bibitem{hu2021lora}
Edward~J Hu et~al.
\newblock Lora: Low-rank adaptation of large language models.
\newblock {\em arXiv preprint arXiv:2106.09685}, 2021.

\bibitem{Goodfellow-et-al-2016}
Ian Goodfellow, Yoshua Bengio, and Aaron Courville.
\newblock {\em Deep Learning}.
\newblock MIT Press, 2016.
\newblock \url{http://www.deeplearningbook.org}.

\bibitem{kullback1951information}
Solomon Kullback and Richard~A. Leibler.
\newblock On information and sufficiency.
\newblock {\em The Annals of Mathematical Statistics}, 22(1):79--86, 1951.

\bibitem{lin1991divergence}
Jianhua Lin.
\newblock Divergence measures based on the shannon entropy.
\newblock {\em IEEE Transactions on Information Theory}, 37(1):145--151, 1991.

\bibitem{cover2006elements}
Thomas~M. Cover and Joy~A. Thomas.
\newblock {\em Elements of Information Theory}.
\newblock Wiley-Interscience, Hoboken, NJ, 2 edition, 2006.

\bibitem{qwen2024qwen2}
Qwen Team.
\newblock Qwen2 technical report.
\newblock {\em arXiv preprint arXiv:2407.10671}, 2024.

\bibitem{elhage2022superposition}
Nelson Elhage, Neel Nanda, Chris Olah, Nicholas Joseph, Dario Amodei, and Shan
  Carter.
\newblock Toy models of superposition.
\newblock {\em Transformer Circuits Thread}, 2022.
\newblock Anthropic.

\bibitem{wei2022emergent}
Jason Wei, Yi~Tay, Rishi Bommasani, Colin Raffel, Barret Zoph, Sebastian
  Borgeaud, Dani Yogatama, Maarten Bosma, Denny Zhou, Donald Metzler, et~al.
\newblock Emergent abilities of large language models.
\newblock {\em Transactions on Machine Learning Research}, 2022.

\bibitem{schaeffer2023mirage}
Rylan Schaeffer, Brando Miranda, and Sanmi Koyejo.
\newblock Are emergent abilities of large language models a mirage?
\newblock {\em Advances in Neural Information Processing Systems}, 2023.

\bibitem{amodei2016concrete}
Dario Amodei, Chris Olah, Jacob Steinhardt, Paul Christiano, John Schulman, and
  Dan Man{\'e}.
\newblock Concrete problems in ai safety.
\newblock {\em arXiv preprint arXiv:1606.06565}, 2016.

\end{thebibliography}
